# Independence with Lower and Upper Probabilities

**Lonnie Chrisman**
School of Computer Science
Carnegie Mellon University
Pittsburgh, PA 15217
chrisman@cs.cmu.edu

## Abstract

It is shown that the ability of the interval probability representation to capture epistemological independence is severely limited. Two events are epistemologically independent if knowledge of the first event does not alter belief (i.e., probability bounds) about the second. However, independence in this form can only exist in a 2-monotone probability function in degenerate cases — i.e., if the prior bounds are either point probabilities or entirely vacuous. Additional limitations are characterized for other classes of lower probabilities as well. It is argued that these phenomena are a matter of interpretation. They appear to be limitations when one interprets probability bounds as a measure of epistemological indeterminacy (i.e., uncertainty arising from a lack of knowledge), but are exactly as one would expect when probability intervals are interpreted as representations of ontological indeterminacy (indeterminacy introduced by structural approximations).

## 1 Introduction

Let $(\Omega, \mathcal{F})$ be a probability space, and let $\underline{P}, \overline{P} : \mathcal{F} \longrightarrow [0,1]$ be set-functions on this space satisfying the following properties for any $A, B \in \mathcal{F}$ with $A \cap B = \emptyset$:

1. $\underline{P}(\emptyset) = \overline{P}(\emptyset) = 1$, $\underline{P}(\Omega) = \overline{P}(\Omega) = 1$
2. $\underline{P}(A) + \overline{P}(\bar{A}) = 1$
3. $\underline{P}(A) + \underline{P}(B) \leq \underline{P}(A \cup B)$ (Super-additivity)
4. $\overline{P}(A) + \overline{P}(B) \geq \overline{P}(A \cup B)$ (Sub-additivity)

where $\bar{A}$ denotes $\Omega - A$, the *complement* of $A$. Then $\underline{P}$ and $\overline{P}$ are called *lower* and *upper probability functions* respectively. It is always the case that $\underline{P}(A) \leq \overline{P}(A)$. It is only necessary to store one or the other of $\underline{P}$ and $\overline{P}$, since each can be obtained using Property 2 once the other is known. The *lower* and *upper probability envelopes* of a non-empty set of distributions $\mathcal{P}$ on $(\Omega, \mathcal{F})$ are functions

$$\underline{P}(A) = \inf\{P(A) : P \in \mathcal{P}\} \quad (1a)$$
$$\overline{P}(A) = \sup\{P(A) : P \in \mathcal{P}\} \quad (1b)$$

Every lower (upper) probability envelope is a lower (upper) probability. Thus, the lower probability representation provides a convenient description of a set of distributions.

A number of uses have been suggested for lower probabilities, and their use is rapidly increasing. Some feel that the use of a single exact distribution in Bayesian-style inference fails to satisfactorily distinguish between uncertainty and ignorance or between certainty and confidence, and therefore a more general representation such as lower probability functions may be a superior representation of belief [21, 22, 24, 32, 33]. Lower probabilities may also arise from incomplete or partial elicitation, such as when insufficient knowledge is available, or when it is too time consuming to obtain obtain the necessary knowledge to warrant the precision inherent in exact probabilities [2, 16, 18]. Lower probabilities are also useful for studying sensitivity and robustness in probabilistic inference [1, 36, 40], and they can be used to weigh computation effort against modeling precision [9]. They arise in group decision problems [24, 28] and in axiomatic approaches to uncertainty when the axioms of probability are weakened [17, 26, 33, 36]. They arise when determining constraints on probabilities given only the probabilities on a finite set of other events [14, 27]. Finally, they may result from the abstraction of more detailed probabilistic models [5, 6, 19].

This paper examines a particular problem that arises with the use of lower probability functions when we attempt to model independent events. We limit our consideration to Bayesian-style updating of lower probability functions, such that when evidence $E \subset \Omega$ is learned, each distribution in $\mathcal{P}$ is updated according to Bayes's rule, yielding the new updated bounds

$$\underline{P}(A|E) = \inf\{P(A|E) : P \in \mathcal{P}, P(E) > 0\}$$
$$\overline{P}(A|E) = \sup\{P(A|E) : P \in \mathcal{P}, P(E) > 0\} \quad (2)$$

It is often impossible to capture *epistemological independence* within a lower probability representation ([36] uses the term *epistemic independence*. Here we follow [37] by using the term *epistemological*). Two events being epistemologically independent would imply that learning the truth about the first should not alter belief (i.e., probability bounds) on the second. Specifically, if the initial lower probability is 2-monotone (defined later), we show that epistemological independence cannot be captured unless we are in one of the degenerate cases where $\underline{P} = \overline{P}$ (i.e., the point probability case) or $\underline{P} = 0$ and $\overline{P} = 1$ (the vacuous case). We also characterize other circumstances in which a lower probability cannot capture this type of independence.

We argue that the apparent inability of lower probabilities to capture independence is a matter of interpretation. Although we have specified that the bounds arise as extrema of $\mathcal{P}$, we have not specified why a set of distributions should be considered in the first place. The apparent difficulty arises from an implicit assumption that the set of distributions is used to represent some form of *epistemological indeterminacy* — that is, a degree of knowledge or the lack of knowledge about the true situation. The qualitative properties of the lower probability representation, particularly with respect to representing independence, but also in terms of related phenomena such as *dilation* [29], make it poorly matched for epistemologically-based interpretations. We propose instead an alternative interpretation, whereby the bounds arise as a result of *ontological* (i.e., structural) considerations. In our interpretation, (point) probabilities capture the epistemological indeterminacy, but (approximate) structural assumptions placed on a model from above introduce additional indeterminacy with a qualitatively different nature, one in which the behavior of the bounds under conditioning can be logically interpreted.

## 2  Coin Tossing Example

Suppose we have two coins which we consider to be physically independent of each other. We are going to toss both coins and observe their outcomes. Each coin has only two possibilities, $\{heads, tails\}$, and each has its own (unrelated) bias on the probability of landing heads, which we know only to be between 1/4 and 3/4. First, we wish to characterize our knowledge using a lower probability function. We denote the four possible outcomes as $\Omega = \{h_1h_2, h_1t_2, t_1h_2, t_1t_2\}$.

Since it may be the case that both coins have a 1/4 probability of heads, we assign $\underline{P}(\{h_1h_2\}) = 1/16$. Similarly, they may both have a 3/4 probability of coming up heads, so we assign $\overline{P}(\{h_1h_2\}) = 9/16$. Carrying out this logic for all of the 16 possible sets of outcomes, we obtain the bounds in Figure 1.

Let $\mathcal{P}(\underline{P})$ denote the set of all probability distributions consistent with the bounds in Figure 1 — i.e.,

| For the sets: | $\underline{P}$ | $\overline{P}$ |
|---|---|---|
| $\emptyset$ | 0 | 0 |
| $\{h_1h_2\}, \{h_1t_2\}, \{t_1h_2\}$, or $\{t_1t_2\}$ | 1/16 | 9/16 |
| $\{h_1h_2, h_1t_2\}, \{t_1h_2, t_1t_2\}, \{h_1h_2, t_1h_2\}$ or $\{h_1t_2, t_1t_2\}$ | $\frac{1}{4}$ | $\frac{3}{4}$ |
| $\{h_1h_2, t_1t_2\}$ or $\{h_1t_2, t_1h_2\}$ | 3/8 | 5/8 |
| $\{h_1t_2, t_1h_2, t_1t_2\}, \{h_1h_2, t_1h_2, t_1t_2\}$, $\{h_1h_2, h_1t_2, t_1t_2\}$, or $\{h_1h_2, h_1t_2, t_1t_2\}$ | $\frac{7}{16}$ | $\frac{15}{16}$ |
| $\Omega = \{h_1h_2, h_1t_2, h_2t_1, t_1t_2\}$ | 1 | 1 |

Figure 1: Bounds on the possible joint outcomes of two coins.

| For the sets: | $\underline{P}$ | $\overline{P}$ |
|---|---|---|
| $\emptyset, \{t_1h_2\}, \{t_1t_2\}$, or $\{t_1h_2, t_1t_2\}$ | 0 | 0 |
| $\{h_1h_2\}, \{h_1t_2\}, \{h_1h_2, t_1h_2\}$ $\{h_1t_2, t_1t_2\}, \{h_1h_2, t_1t_2\}, \{h_1t_2, t_1h_2\}$ $\{h_1t_2, t_1h_2, t_1t_2\}$, or $\{h_1h_2, t_1h_2, t_1t_2\}$ | $\frac{1}{8}$ | $\frac{7}{8}$ |
| $\{h_1h_2, h_1t_2\}, \{h_1h_2, h_1t_2, t_1h_2\}$, $\{h_1h_2, h_1t_2, t_1t_2\}$, or $\Omega = \{h_1h_2, h_1t_2, t_1h_2, t_1t_2\}$ | 1 | 1 |

Figure 2: Bounds after conditioning on $H_1 = \{h_1h_2, h_1t_2\}$.

$P \in \mathcal{P}(\underline{P})$ if and only if $\underline{P}(A) \leq P(A) \leq \overline{P}(A)$ for all $A \subset \Omega$.

Suppose we observe the outcome of the first coin to be *heads* without observing the outcome of the second coin. This gives us the conditioning event $H_1 = \{h_1h_2, h_1t_2\}$. We then update our bounds given the new evidence as follows:

$$\underline{P}(A|H_1) = \inf\{P(A|H_1) : P \in \mathcal{P}(\underline{P}), P(H_1) > 0\}$$
$$\overline{P}(A|H_1) = \sup\{P(A|H_1) : P \in \mathcal{P}(\underline{P}), P(H_1) > 0\} \quad (3)$$

This yields the new bounds shown in Figure 2. Notice the new bounds for the event $H_2 = \{h_1h_2, t_1h_2\}$, which were previously [1/4, 3/4], but are now [1/8, 7/8]. The outcomes of the two coins are supposedly independent, yet learning the outcome of the first coin had a marked influence on our beliefs about the outcome of the second (independent) coin. The representation has clearly failed to capture the independence.

The inability to capture this independence is related to the fact that $\mathcal{P}(\underline{P})$ includes distributions in which the coins are not independent. Using the set

$$\mathcal{P} = \{P(\{x_1x_2\}) = P_1(\{x_1\})P_2(\{x_2\}) : x \text{ stands for}$$
$$h \text{ or } t, 1/4 \leq P_i(\{h_1\}) \leq 3/4, i = 1, 2\} \quad (4)$$

would more accurately reflect the complete knowledge in this example. This is what [36] calls the sensitivity analysis approach to independence, and [11] call type-1 independence. This (non-convex) set of probabilities is shown graphically in Figure 3. For this set, $\underline{P}(H_2)$, as defined by (1) and shown in Figure 1, is equal to $\underline{P}(H_2|H_1)$, as defined by (2), so a perfect representation of $\mathcal{P}$ would capture the independence. However, a primary reason for studying the lower probability representation is for the purpose of using it as a complete representation of belief. Because independence plays a central role in many theories of subjective belief, the fact that the lower probability representation has problems capturing independence is significant. In the remainder of the paper, we will characterize this (in)ability to capture independence and examine what this suggests for the interpretation of the representation.

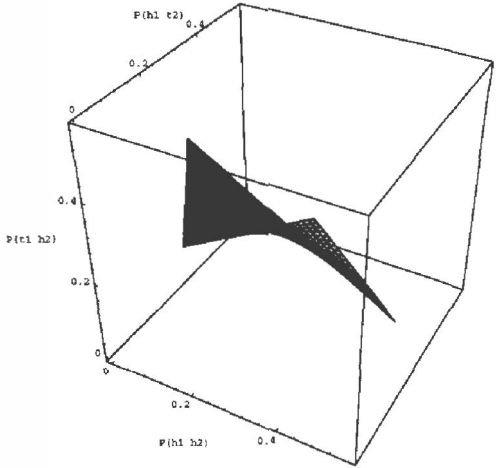

Figure 3: The set $\mathcal{P}$ shown graphically. Because the probabilities of the four outcomes must sum to one in each probability distribution, we can plot each distribution as a point in three dimensions. The set $\mathcal{P}$ resembles a piece twisted of paper. It is nonconvex — for example, $\langle 1/16, 3/16, 3/16, 9/16 \rangle$ and $\langle 9/16, 3/16, 3/16, 1/16 \rangle$ are in $\mathcal{P}$, but their average, $\langle 5/16, 3/16, 3/16, 5/16 \rangle$ is not.

## 3  Lower Probability

Before examining the independence issues in more detail, this section defines some notation and reviews some of the basics of the lower probability representation. The subsequent section characterizes independence issues. The properties and terminology in this section has been developed and utilized by [3, 4, 7, 36, 37] and others.

A lower probability is a function obeying the properties listed in the introduction. A probability distribution, $P$, is *consistent* with a lower probability $\underline{P}$ if for every $A \in \mathcal{F}, \underline{P}(A) \leq P(A)$. We denote by $\mathcal{P}(\underline{P})$ the set of all distributions consistent with $\underline{P}$. The conditions in the introduction are not strong enough to ensure that $\mathcal{P}(\underline{P}) \neq \emptyset$. A lower probability $\underline{P_1}$ *dominates* $\underline{P_2}$ if for all $A \in \mathcal{F}$, $\underline{P_1}(A) \geq \underline{P_2}(A)$, in which case $\mathcal{P}(\underline{P_1}) \subset \mathcal{P}(\underline{P_2})$.

Denote the lower probability envelope $\underline{P}$ obtained from $\mathcal{P}$ using (1a) by $\underline{P}[\mathcal{P}]$. It does not follow that $\mathcal{P}(\underline{P}[\mathcal{P}]) = \mathcal{P}$. In other words, many different sets of distributions share the same bounds. The set $\mathcal{P}(\underline{P}[\mathcal{P}])$ is called the *majorization* of $\mathcal{P}$. When $\mathcal{P}(\underline{P}[\mathcal{P}]) = \mathcal{P}$, $\mathcal{P}$ is said to be *closed to majorization* ([39]). When $\underline{P} = \overline{P}$, then $\underline{P}$ is a probability distribution.

Every lower probability function is *monotone* (sometimes called *1-monotone*), meaning that $\underline{P}(A) \leq \underline{P}(B)$ whenever $A \subset B$. A stronger property called 2-monotonicity is often useful. A lower probability $\underline{P}$ is *2-monotone* when for every $A, B \in \mathcal{F}$,

$$\underline{P}(A) + \underline{P}(B) \leq \underline{P}(A \cap B) + \underline{P}(A \cup B)$$

Two-monotonicity is a sufficient (but not necessary) condition to ensure that $\underline{P}$ is a lower envelope. Two-monotonicity is usually necessary for obtaining exact closed-form manipulation formula, and is therefore usually assumed in practice.

The lower and upper probabilities conditioned on event $E \in \mathcal{F}$ are given by

$$\underline{P}(A|E) = \inf\{P(A|E) : P \in \mathcal{P}(\underline{P}), P(E) > 0\}$$
$$\overline{P}(A|E) = \sup\{P(A|E) : P \in \mathcal{P}(\underline{P}), P(E) > 0\}$$

It is well-known [7, 13, 15, 38, 39] that when $\underline{P}$ is 2-monotone,

$$\underline{P}(A|E) = \frac{\underline{P}(A \cap E)}{\underline{P}(A \cap E) + \overline{P}(\bar{A} \cap E)}$$

$$\overline{P}(A|E) = \frac{\overline{P}(A \cap E)}{\overline{P}(A \cap E) + \underline{P}(\bar{A} \cap E)} \quad (5)$$

whenever $\underline{P}(E) > 0$, and $\underline{P}(A|E) = 1$ whenever $A \subset E$ and $\overline{P}(E) > \underline{P}(E) = 0$. If $\overline{P}(E) = 0$, then the conditional lower probability is undefined.

Let $\Omega$ be finite, $\mathcal{F} = 2^\Omega$. The *Möbius transform* of $\underline{P}$ is a set function $m : \mathcal{F} \longrightarrow \Re$ defined by ([30, pg. 39])

$$m(A) = \sum_{B \subset A} (-1)^{|A-B|} \underline{P}(B)$$

If $m(A) \geq 0$ for all $A \in \mathcal{F}$, then $\underline{P}$ is said to be a *belief function*. Belief functions are used by the Dempster-Shafer theory ([30]) and in the Transferable Belief Model ([32]), but in those theories are given evidential interpretations rather than the lower probabilistic interpretation of interest here (see [20] and [31] for good discriptions of the difference). Belief functions are also what [4] terms *infinitely-monotone capacities* ([34]). Every infinitely-monotone lower probability (i.e., belief function) is also 2-monotone, but the converse does not hold.

The Möbius transform is information preserving, so that the original function can be recovered from $m$

using the *inverse Möbius transform*, given by

$$\underline{P}(A) = \sum_{B \subset A} m(B)$$

Subsets $A \in \mathcal{F}$ with $m(A) \neq 0$ are called the *focal elements* of $\underline{P}$.

Let $\underline{P_A}$ and $\underline{P_B}$ be lower probabilities on $\Omega_A$ and $\Omega_B$ respectively. The *meta-Markov combination* of $\underline{P_A}$ and $\underline{P_B}$ ([10]), denoted $\mathcal{P} = \mathcal{P}(\underline{P_A}) \otimes \mathcal{P}(\underline{P_B})$, is the set of distributions on $\Omega = \Omega_A \times \Omega_B$ given by

$$\mathcal{P} = \left\{ P : \begin{array}{c} P(\{ab\}) = P_A(\{a\}) P_B(\{b\}), \\ P_A \in \mathcal{P}(\underline{P_A}), P_B \in \mathcal{P}(\underline{P_B}) \end{array} \right\}$$

This is the set consisting of all independent events, i.e., the set shown in Figure 3 for the coin tossing example. We also write $\underline{P} = \underline{P_A} \otimes \underline{P_B}$ for the majorization of this set.

## 4  Representation of Independence

Probabilistic independence of $A$ and $B$ is characterized by either of two properties:

1. $P(A \cap B) = P(A) P(B)$
2. $P(A|B) = P(A)$ when $P(B) > 0$

In the case of a probability distribution each of these imply the other. It is not hard to see that in the case of lower probabilities the two properties do not imply each other, and therefore it seems natural to define independence for lower probabilities as the conjunction of both properties, i.e., $A$ and $B$ are independent whenever

1. $\underline{P}(A|B) = \underline{P}(A)$ and $\overline{P}(A|B) = \overline{P}(A)$,
   when $\overline{P}(B) > 0$                    (irrelevance)
2. $\underline{P}(A \cap B) = \underline{P}(A)\underline{P}(B)$ and
   $\overline{P}(A \cap B) = \overline{P}(A)\overline{P}(B)$    (factorization)

In fact, [37] give exactly this definition. However, as the subsequent theorems show, the two are often mutually incompatible.

**Theorem 1** *When $\underline{P}$ is 2-monotone, the following conditions cannot all hold:*

1. $\underline{P}(A) = \underline{P}(A|B)$
2. $\underline{P}(A \cap B) = \underline{P}(A)\underline{P}(B)$ and
   $\overline{P}(\bar{A} \cap B) = \overline{P}(\bar{A})\overline{P}(B)$.
3. $0 < \underline{P}(A) < 1$
4. $\overline{P}(B) > \underline{P}(B)$,

*where $\underline{P}(A|B) = \inf\{P(A|B) : P \in \mathcal{P}(\underline{P}), P(B) > 0\}$ and $A, B \subset \Omega$. If any one of the four properties is removed, the three remaining properties can co-exist.*

The above shows that a 2-monotone lower probability cannot exhibit the desired properties of independence except in the degenerate cases where $\underline{P}(A) = 0$ or $\underline{P}(A) = 1$, i.e., when $\underline{P}$ is entirely uninformative (vacuous) about $A$, or when $\underline{P}(B) = \overline{P}(B)$ (i.e., $\underline{P}(B)$ is a point probability on $B$).

We can demonstrate the applicability of the above theorem on the following example.

**Example 1:** [The Extended Monte Hall Problem] Jane is a contestant on *Let's Make a Deal*, a game show. Presented with four curtains, behind only one of which is a prize, she selects Curtain 1. The host then reveals *first* that there is nothing behind Curtain 4, and *second* that there is nothing behind Curtain 3. Making the assumptions that initially the location of the prize is equiprobable, that the host will always show two empty unselected curtains, that the unselected curtain not revealed is chosen uniformly, that the curtain order is independent of all other aspects of the problem, and that her knowledge about how the curtain order is picked is characterized by a vacuous lower probability, what is the lower probability of winning if she does not change her selection? What is the lower probability of winning if she does change selection? Assume Jane initially captures all the knowledge of the problem using only a lower probability distribution, and that the frame of discernment used includes the curtain order (so that $\Omega = \{123, 132, 124, 142, 134, 143, 234, 243, 324, 342, 423, 432\}$ where $ijk$ abbreviates "the prize is behind $i$, the host reveals first $j$ and second $k$).

The information stated above is encoded in the lower probability function with the Möbius transform focal elements

$$m(\{123, 132\}) = m(\{124, 142\}) = m(\{134, 143\}) = 1/12$$
$$m(\{234, 243\}) = m(\{324, 342\}) = m(\{423, 432\}) = 1/4 \tag{6}$$

The initial lower probability is in this case a belief function, and therefore 2-monotone. Conditions 1, 2, 4, and 5 of Theorem 1 are satisfied, so by observing the order in which the curtains are revealed, we know her lower probability for each of the two questions is effected. In particular, after observing the revealed curtains and their order, her remaining belief becomes vacuous, i.e.,

$$\underline{P}(\{134, 143\}|\{143, 243\}) = \underline{P}(\{234, 243\}|\{143, 243\}) = 0$$
$$\overline{P}(\{134, 143\}|\{143, 243\}) = \overline{P}(\{234, 243\}|\{143, 243\}) = 1$$

Had she ignored curtain order entirely, using $\Omega = \{12, 13, 14, 21, 31, 41\}$, where $ij$ abbreviates "the prize is behind $i$, the host does not reveal $j$," her final beliefs (according to Bayes's rule) are that she'd have a point-probability of 3/4 of winning if she changes her selection, or a point-probability of 1/4 if she does not change. Once again, the result should be completely independent of the curtain order, but in the lower probability representation the influence can be quite dramatic. □

Theorem 1 covers a wide class of lower probabilities that are of great interest. However, there are in addition lower probability functions that are not even 2-monotone, but for which independence properties cannot hold. The lower probability in the initial coin tossing example (Figure 1) was such an example — it is not 2-monotone. Therefore, it is possible to obtain further characterizations for when the independence properties cannot co-exist. The following characterization covers the coin tossing example.

**Theorem 2** *Suppose $\mathcal{P}(\underline{P})$ is the set of distributions consistent with $\underline{P}$, and let $A, B \subset \Omega$. If there exists a $P \in \mathcal{P}(\underline{P})$ such that $P(A) = \underline{P}(A)$, $P(B) > 0$ and $P(A) > P(A|B)$, then $\underline{P}(A) > \underline{P}(A|B)$.*

*Dually, if there exists a $P \in \mathcal{P}(\underline{P})$ such that $P(A) = \overline{P}(A)$ and $P(A) < P(A|B)$, then $\overline{P}(A) < \overline{P}(A|B)$.*

That Theorem 2 covers the initial coin tossing example is immediately seen with the distribution $\langle 1/16, 3/8, 3/16, 3/8 \rangle$, which is consistent with the bounds in Figure 1 and satisfies the conditions in Theorem 2.

Theorem 2 is closely related to [29, Theorems 2 and 3] which state virtually identical conditions under which $B$ will *dilate* the lower probability bounds (i.e., the posterior bounds after updating on $B$ will strictly contain the prior bounds). Clearly, if the bounds dilate on an event that is supposed to be independent, the lower probability is not exhibiting the independence properties. However, the connection between dilation and independence is actually closer than this. For example, [29, Theorem 1] shows that dilation can only occur if the set of distributions with the desired independence property intersects the set of consistent distributions. The following theorem emphasizes this connection between the independence properties and dilation — independent events cannot cause a set of independent distributions to contract.

**Theorem 3** *Let $\underline{P} = \underline{P}_{\mathbf{A}} \otimes \underline{P}_{\mathbf{B}}$, with $\underline{P}(A|B)$ given by (2). (Similarly for $\overline{P}$). Then for $A = A' \times \Omega_{\mathbf{B}}$ and $B = \Omega_{\mathbf{A}} \times B'$, where $A' \subset \Omega_{\mathbf{A}}$ and $B' \subset \Omega_{\mathbf{B}}$:*

1. $\underline{P}(A \cap B) = \underline{P}(A)\underline{P}(B)$
2. $\overline{P}(A \cap B) = \overline{P}(A)\overline{P}(B)$
3. $\underline{P}(A|B) \leq \underline{P}(A) \leq \overline{P}(A) \leq \overline{P}(A|B)$

In addition to the connection with dilation (Items 3 and 4), Items 1 and 2 of Theorem 3 demonstrate that the factorization property of independence is always a property of lower probabilities when we are dealing with independent events. Recall that these conditions appeared in Theorem 1.

The idea that information about one fact should not influence beliefs regarding certain other facts is an important component in many formalizations of knowledge representation. The theorems in this section demonstrate that the lower probability representation often cannot exhibit this property except in degenerate cases.

An alternative version of epistemological independence is possible. Instead of requiring that independent events do not affect conditioned probability bounds, events can be called independent whenever $\underline{P}(A|B) \leq \underline{P}(B)$. This version is identified by [12, Definition 4.3] with the rationale that independent events should not contribute additional information, a requirement that is much weaker than the irrelevance requirement. This weaker requirement is compatible with the factorization property and, as evidenced by the results of this section, is a preferable property for independence within the lower probability framework. It also should be noted that while [37] define independence as having both properties hold, [36] defines epistemological independence as the first property (irrelevance) only.

## 5 Abstraction

This section examines factorization and the relationship between a factored lower probability and its consistent probability distributions. This relationship is central to the interpretation of lower probability considered in the subsequent section.

Let $P^*$ be an arbitrary probability distribution on $\Omega = \Omega_{\mathbf{A}} \times \Omega_{\mathbf{B}}$. Specifically, it is *not* necessarily the case that $\mathbf{A} \perp\!\!\!\perp \mathbf{B}[P^*]$ (that $\mathbf{A}$ is independent of $\mathbf{B}$ with respect to $P^*$).

**Definition 1** *A lower probability $\underline{P}$ on $\Omega$ is an abstraction of $P^*$ relative to the assertion $\mathbf{A} \perp\!\!\!\perp \mathbf{B}$ when*

1. $P^* \in \mathcal{P}(\underline{P})$
2. $\underline{P} = \underline{P}_{\mathbf{A}} \otimes \underline{P}_{\mathbf{B}}$

*where $\underline{P}_{\mathbf{A}}$ is a lower probability on $\Omega_{\mathbf{A}}$, $\underline{P}_{\mathbf{B}}$ is a lower probability on $\Omega_{\mathbf{B}}$. $\underline{P}$ is a* proper abstraction *if it is not dominated by any other abstraction of $P^*$ relative to $\mathbf{A} \perp\!\!\!\perp \mathbf{B}$.*

It is worth emphasizing that an abstraction is factorizable (Item 2) and captures information about $P^*$ without introducing information that is not implied by $P^*$. No abstraction can capture strictly more information than a proper abstraction without introducing information that is not implied by $P^*$; however, a proper abstraction is *not* unique — there may be an arbitrarily large number of proper abstractions relative to a single independence assertion, and each of these may contain information not contained by the others. Note that by definition, any abstraction is closed to majorization.

Theorem 3 has already revealed that $\underline{P}(A \cap B) = \underline{P}(A)\underline{P}(B)$. This does not, however, describe the lower probability of *non-rectangular sets* (those which cannot be written as $A \times B$). The full characterization of all sets is most conveniently stated in terms of the Möbius transform.

**Theorem 4** *If $\underline{P}$ is an abstraction of a distribution $P$ relative to $\mathbf{A} \perp\!\!\!\perp \mathbf{B}$, and $m$ is the Möbius transform of $\underline{P}$, then*

$$m(X) = \begin{cases} m_{\mathbf{A}}(A) m_{\mathbf{B}}(B) & \text{when } \begin{array}{l} X = A \times B, \\ A \subset \Omega_{\mathbf{A}}, B \subset \Omega_{\mathbf{B}} \end{array} \\ 0 & \text{otherwise} \end{cases}$$

*where $m_{\mathbf{A}}$ and $m_{\mathbf{B}}$ are the Möbius transforms of $\underline{P_{\mathbf{A}}}$ and $\underline{P_{\mathbf{B}}}$, $\underline{P_{\mathbf{A}}}(A) = \underline{P}(A \times \Omega_{\mathbf{B}})$, and $\underline{P_{\mathbf{B}}}(B) = \underline{P}(\Omega_{\mathbf{A}} \times B)$.*

Theorem 4 does not require the abstraction to be proper.

It is possible to generalize the concept of an abstraction relative to a single independence assertion to the concept of an abstraction relative to a set of conditional independence assertions. This introduces a number of complications beyond the scope of the current paper. A general concept of factorization (decomposition) of lower probabilities is developed in [8].

The concept of a proper abstraction immediately suggests an interpretation for probability bounds — namely, that a lower probability is an abstraction of some (more detailed) probability distribution. The exact identity of this distribution is lost — it is known only to be in $\mathcal{P}(\underline{P})$. The next section develops this interpretation.

## 6 The Ontological Interpretation

This section introduces an interpretation of lower probability. This interpretation resolves many of the apparent limitations discussed above, and provides an interpretation that suggests important uses for lower probabilities.

Let us assume that a probability or a lower probability function is to serve as a model of some system or phenomena, as is often the case. Models are by their very nature approximations or abstractions of the actual system being modeled, and as such they bring with them a certain amount of indeterminacy. By including probabilities or lower probabilities in the description of the model, we often aim to quantify this indeterminacy explicitly.

Constructing a model of a system involves two basic steps: (1) Choosing an ontology, and (2) Filling in the knowledge required by the ontology. An *ontology* specifies the language used to describe the system, as well as structural and parametric assumptions that are built into the model. We can think of an ontology as identifying a set of parameters that must be filled in to specify the actual knowledge of the particular system being modeled, as well as a set of variables that are used to describe particular problem instances. The ontology (alone) leaves the values of the parameters unspecified, for this is the epistemological information. Once the parameters are specified, the model is completely specified, and the ontology relates these parameters to each other and to the problem instance variables. We refer to these two levels as the *ontological level* and the *epistemological level*.

In the case where two coins are tossed, choices at the ontological level include assuming that exactly one of only two possible outcomes can occur for each coin, that the outcome of each coin can be characterized by a single probability, that outcomes of consecutive tosses are independent of one another and of the other coin. These correspond to choices of language, parametric assumptions, and structural assumptions respectively. This ontology requires two parameters to be filled in to completely specify the model. The values for the two coins' biases are the knowledge at the epistemological level.

Indeterminacy in a model can arise at either level, and we refer to these as *ontological indeterminacy* or *epistemological indeterminacy* (these terms were coined by [37]). However, ontological indeterminacy can only exist when there is epistemological indeterminacy, because otherwise our model is nothing more than an exact description of the true situation.

Probability provides a very good representation for epistemological indeterminacy. We argue, therefore, that (non-point) lower probabilities are inappropriate for quantifying pure epistemological indeterminacy. This viewpoint is much along the lines of a strict "Bayesian" interpretation of probability, and in stark contrast to epistemological interpretations of probability bounds offered by [18, 20, 21, 23, 24, 25], and others, in which imprecision arises from a deficiency of knowledge or training data. Under our proposed interpretation, interval probability bounds arise *only* as a result of ontological indeterminacy, i.e., structural assertions that are only approximately true. Thus, when given a lower probability function, we immediately interpret non-point intervals as a reflection of ontological indeterminacy, and probabilities as a reflection of epistemological indeterminacy.

The relationship between epistemological and ontological indeterminacy can be visualized as follows. The epistemological indeterminacy of a rational, coherent agent is quanitified by a probability distribution $P^*$. $P^*$ can be thought of as the agent's deepest beliefs, but these might not be easily accessible to a resource-bounded agent. Inferences are performed using a model that includes ontological assumptions convenient for the problem(s) being solved. The model used by the agent is an abstraction of $P^*$ relative to the ontology's independence assertions. Ideally it is a proper abstraction so that a minimal amount of additional in-determinacy is introduced by the abstraction.

It has been said that "the assumption of conditional independence is usually false" [35]. By asserting a conditional independence assertion, an agent is more typically asserting a belief that two events are *almost* conditionally independent given a third. An agent might assume, for example, that gravitational acceleration is independent of an object's height because it results in a useful model, even though deep down at the epistemological level the agent does not believe they are truly independent. The result of this structural approximation is that ontological indeterminacy is introduced.

### 6.1 Coin Tossing Revisited

It is instructive to apply this interpretation to the coin tossing example considered earlier. Consider the probability distribution, $P^*$, given by

$$P^*(\{h_1h_2\}) = P^*(\{t_1t_2\}) = 7/16$$
$$P^*(\{t_1h_2\}) = P^*(\{h_1t_2\}) = 1/16 \quad (7)$$

This probability distribution quantifies epistemological indeterminacy. Since there is no structure (i.e., no independence or parametric restriction), there is no ontological indeterminacy, so the point probability on the joint space captures all the agent's uncertainty. If a rational agent had unlimited time and resources to access and compute the ramifications of its deepest beliefs, $P^*$ is the full assessment of beliefs it would obtain.

However, suppose the agent models the coins as independent. From (7), it is clear the agent does not really believe the coins to be independent — this is a structural approximation. There are several possible rationale for the agent imposing this artificial structure on its model: to simplify (factorize) computation, to reduce the number of parameters that must be assessed, to obtain a structure that is better suited for explanation, to reason at different hierarchical levels of abstraction, etc.

The agent adopts (or subjectively estimates) a proper abstraction of $P^*$ relative to this independence assertion. An infinite number of proper abstractions are possible, one of which is obtained by setting $\underline{P}(H_1) = \underline{P}(H_2) = \underline{P}(T_1) = \underline{P}(T_2) = 1/4$, which is shown in Figure 4. The lower probability in Figure 1 contains Möbius assignments on two non-rectangular sets, but is otherwise comparable to the lower probability of Figure 4. In [12], de Campos and Huete call Figure 4 a type-2 product, and Figure 1 a type-1 product, and relate the two with their Proposition 3.6.

When inference is performed using $\underline{P}$, one should not assume anything about $P^*$ except what is implied as a result of $\underline{P}$ being an abstraction of $P^*$. So, for example, $\underline{P}$ is also a proper abstraction of the distribution

$$P(\{h_1h_2\}) = P(\{t_1t_2\}) = 1/16$$
$$P(\{t_1h_2\}) = P(\{h_1t_2\}) = 7/16 \quad (8)$$

| For the sets: | $\underline{P}$ | $\bar{P}$ |
|---|---|---|
| $\emptyset$ | 0 | 0 |
| $\{h_1h_2\}, \{h_1t_2\}, \{t_1h_2\}$ or $\{t_1t_2\}$ | 1/16 | 9/16 |
| $\{h_1h_2, h_1t_2\}, \{t_1h_2, t_1t_2\}, \{h_1h_2, t_1h_2\}$ or $\{h_1t_2, t_1t_2\}$ | 1/4 | 3/4 |
| $\{h_1h_2, t_1t_2\}$ or $\{h_1t_2, t_1h_2\}$ | 1/8 | 7/8 |
| $\{h_1t_2, t_1h_2, t_1t_2\}, \{h_1h_2, t_1h_2, t_1t_2\},$ $\{h_1h_2, h_1t_2, t_1t_2\}$ or $\{h_1h_2, h_1t_2, t_1h_2\}$ | $\frac{7}{16}$ | $\frac{15}{16}$ |
| $\Omega = \{h_1h_2, h_1t_2, t_1h_2, t_1t_2\}$ | 1 | 1 |

Figure 4: Bounds encoding ontological indeterminacy for two independent coins.

Any inference from $\underline{P}$ should be valid for $P$ as well as for $P^*$.

Suppose the agent observes the outcome of the first coin to be heads, without observing the outcome of the second coin. $\underline{P}(H_2|H_1)$ should bound the conditional probability $P(H_2|H_1)$ for any more detailed probabilistic model, and the bound must be valid for *any* consistent distribution. For example, $P^*(H_2|H_1) = 7/8$, but if $P$ is the distribution of (8) then $P(H_2|H_1) = 1/8$. In full, the desired conditional lower probability is indeed that given by (3) and shown in Figure 2. In other words, under the ontological interpretation of lower probability, the bounds that previously seemed to present a paradox are in fact the desired conditional bounds. These new bounds are guaranteed to be consistent with any (ontologically) more detailed model.

The apparent paradox with the coin tossing example of Section 2 only appears paradoxical because of an implicit assumption that the lower probabilities are representing a form of epistemological indeterminacy. By interpreting the intervals of a lower probability as representing ontological indeterminacy, the results of conditioning are precisely what we would expect and desire.

### 6.2 Monte Hall Revisited

In the Monte Hall example, observing the order in which curtains are revealed causes Jane's belief about the prize's location to change from a point probability to total ignorance. This occurs despite the fact that curtain order is modeled as independent of the prize's location. This result, however, is quite reasonable when the lower probability is given an ontological interpretation. We must assume that independence between curtain order and prize location is imposed in order to factorize the lower probability. Furthermore, they are not independent at the epistemological level, for if they were, the belief would be characterized by point bounds.

The lower probability of (6) is an abstraction of a more refined model in which the host encodes the exact location of the prize with the selection of curtain order.

For example, if the prize is behind the lower numbered unrevealed curtain, the curtains opened are revealed with the lowest numbered revealed first. This more refined model is certainly consistent with (6), as is the one where the encoding is reversed.

By adopting the beliefs in (6), Jane must believe that deep down, given enough time and thought, she can figure out how the host encodes the prize's location. The vacuous bounds simply indicate that the use of a more detailed model is certainly warranted for this problem. The assertion that the identity of the curtain not revealed is independent of the order the curtains are revealed hardly an approximation — it is blatantly false — and as a result, vacuous bounds result.

## 7  Conclusion

Lower probabilities are utilized for a great number of purposes within the robust statistics and uncertain inference communities.

However, the results here demonstrate that the representation has significant limitations in its ability to represent epistemological independence, the idea that knowledge of one event should not influence belief about a second event. Theorem 1 showed that independence of this type can never be represented by a 2-monotone lower probability unless the bounds are tight (i.e., a point probability), or the bounds are totally vacuous. Theorem 2 shows that this limitation extends to an even wider class of lower probabilities, and Figure 3 suggests the limitations extend even to more general representations of convex sets of probability distributions.

These limitations almost appear to be paradoxical. However, they are only paradoxes when one interprets probability bounds as an indication of epistemological indeterminacy. For example, one often sees it said that lower probabilities are useful because point probabilities require more precision than available knowledge warrants. The results deal a blow to epistemological interpretations such as this. When lower probability is appropriately interpreted, these limitations and the unusual influence of independent events on probability bounds is entirely natural and fully consistent with the interpretation. The ontological interpretation says simply that epistemological indetermancy (uncertainty due to lack of total knowledge) is appropriately represented by a pure probability distribution. When structural approximations are asserted, ontological indeterminacy is introduced. The lower probability representation captures this ontological indeterminacy.

Several other concepts of independence for lower and upper probabilities, as well as for more general sets of probabilities, are also possible ([11, 12]). There are also several possible types of products that can be formed from marginal lower probability representations, and these result in various relationships between independence concepts and product formula. These relationships are studied in [12] and [36, Section 9.3].

In some cases it may be appropriate for an agent to fully assess its epistemological indeterminacy, thus obtaining a probability distribution $P^*$, and then only later abstract this to a lower probability relative to a more structured or simplified model. This form of hierarchical reasoning can reduce the computational effort for solving specific inferences considerably. Furthermore, for any given inference, the bounds obtained give a quantitative indication of how much precision was lost by using the abstract model, and this in turn gives an indication of whether an answer from the current level of abstraction is sufficient. However, fully assessing epstemological indeterminacy first is not entirely necessary. It is also conceivable that bounds themselves are subjectively estimated without first estimating $P^*$, perhaps by considering only the most extreme situations that violate structural approximations. A precise interpretation is important when making such subjective assessments, since it provides a conceptual basis for chosing specific bounds.

Concepts of independence are central to probabilistic reasoning, and are especially important when it comes to scaling to large domains. A thorough understanding of independence and how it can be properly utilized is equally important to lower probabilistic reasoning. The ontological interpretation may provide a useful foundation for utilizing abstraction and structural approximation in the context of probabilistic inference, ideas that are also important when scaling probabilistic inference to very large domains.

## Acknowledgements

I would like to thank Fabio Cozman for comments on a draft of this paper. The author was supported by NASA-Jet Propulsion Lab Grant No. NGT-51039. The views and conclusions contained within are those of the author and should not be interpreted as representing the official policies, either expressed or implied, of the U.S. government.